\documentclass[a4paper,11pt]{article}

\usepackage[utf8]{inputenc}
\usepackage[T1]{fontenc}
\usepackage{geometry}
\usepackage{amsmath,amssymb,amsfonts}
\usepackage{graphicx}
\usepackage{booktabs}
\usepackage{array}
\usepackage{multirow}
\usepackage{float}
\usepackage{threeparttable}
\usepackage{subcaption}
\usepackage{listings}
\usepackage{mathrsfs}
\usepackage{url}
\usepackage[numbers,sort&compress]{natbib}
\usepackage{hyperref}

\graphicspath{{./figs/}}

\begin{document}

\title{\bfseries Multi-Modal Agents for Power Distribution Defect Detection: An Evaluation of Foundation Models}

\author{Quan Quan%
}

\date{}

\maketitle

\begin{abstract}
The power distribution network is critical to reliable electricity delivery, yet traditional inspection methods face limitations in semantic understanding, generalization, and closed-loop automation. To address these challenges, this paper proposes a Multi-Modal Agent framework specifically for power distribution defect detection. Central to this study is the systematic evaluation of multimodal foundation models as unified cognitive engines. We rigorously assess their integrated performance across three critical capabilities: (1) Perception, where the model must accurately identify equipment and generate expert-level descriptions of defects; (2) Reasoning, where the model interprets visual findings to diagnose causes, assess severity, and plan maintenance strategies based on domain knowledge; and (3) Tool Usage, where the model acts as an autonomous operator to execute actions---such as querying knowledge bases or generating work orders---to achieve closed-loop maintenance. To support this evaluation, a domain-specific evaluation dataset and a comprehensive benchmark are developed. Experimental results demonstrate the strengths and limitations of current foundation models in these three dimensions, providing empirical evidence for deploying autonomous agents in high-stakes industrial environments.
\end{abstract}


\vspace{0.5cm}

\section{Introduction}

As the ``last mile'' of the power grid that directly serves end-users, the power distribution network plays a pivotal role in ensuring the safe and stable operation of modern society. Traditional inspection methods rely heavily on manual labor, which is not only inefficient and labor-intensive but also prone to high operational risks. While deep learning techniques---particularly convolutional neural networks have improved inspection efficiency, their deployment in real-world operation and maintenance scenarios still faces critical bottlenecks, such as the lack of semantic understanding (Semantic Gap), difficulty in recognizing unseen defects (Limited Generalization), and the inability to autonomously trigger workflows (Information Silo).

\begin{figure}[t]
    \centering
    \includegraphics[width=\linewidth]{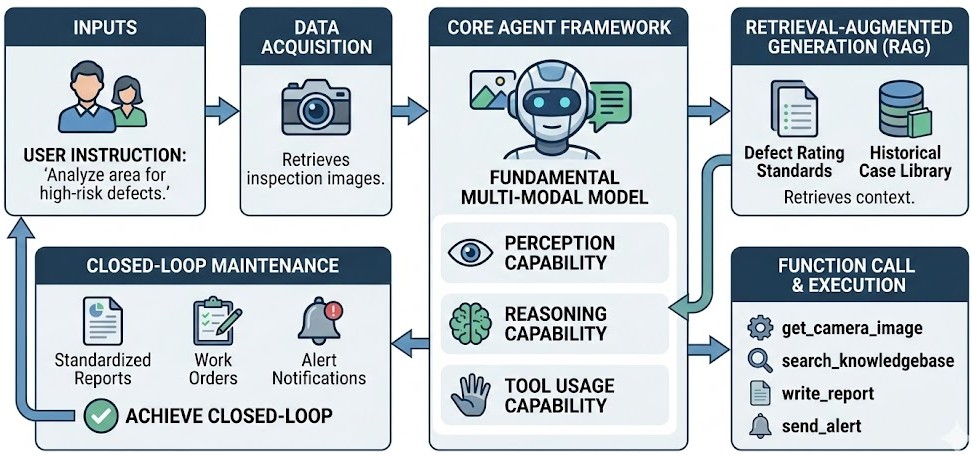}
    \caption{Overview of multi-modal agent framework for power inspection}
    \label{fig:overview}
\end{figure}

To address these challenges systematically, we envision a more integrated paradigm---the Multi-Modal Agent---as a promising pathway toward truly autonomous inspection. Unlike traditional isolated detection models, such an agent is a computational entity capable of independently executing complex inspection tasks. The effectiveness of this agent is critically dependent on its core multimodal foundation model (e.g., integrated VLM-LLM architectures). In this paper, we evaluate these foundation models as unified cognitive engines that must simultaneously demonstrate three essential capabilities to close the loop from detection to maintenance:
\begin{itemize}
    \item (1) Perception: The ability to not only localize objects but also understand and generate detailed, human-like descriptions of complex visual information, effectively bridging the semantic gap between raw images and expert interpretation.
    \item (2) Reasoning: The capacity to act as a ``brain'' that interprets visual findings, applies domain knowledge to diagnose defect causes, assesses severity levels, and formulates logical maintenance plans.
    \item (3) Tool Usage: The capability to translate decisions into executable actions by autonomously invoking external tools---such as querying a knowledge base, retrieving camera feeds, or generating standardized work orders, thereby overcoming the information silo.
\end{itemize}

Despite the rapid development of general-purpose multi-modal models, their application to the power distribution domain---characterized by stringent reliability requirements, domain-specific expertise, and high safety standards---remains underexplored. In particular, there is currently no systematic framework to rigorously evaluate how these \textbf{foundation models} perform across the specific dimensions of perception, reasoning, and tool usage in an industrial context.

To bridge this gap, we implement a comprehensive evaluation framework for the foundation models of multi-modal agents in the context of autonomous power inspection. Our main contributions are as follows:
\begin{itemize}
    \item We propose an evaluation framework for multimodal agents that focuses on assessing their Perception, Reasoning, and Tool Usage capabilities in power inspection tasks, enabling targeted and reproducible performance comparisons across different technology configurations.
    \item We develop a multi-task, multi-dimensional benchmark tailored to power distribution scenarios. The benchmark includes a multimodal dataset for evaluating perception and reasoning, as well as a suite of complex inspection scenarios designed to test the agent's end-to-end tool execution and task completion capability.
    \item Using the proposed framework, we systematically evaluate different agent architectures and core models in power inspection tasks. Our results reveal both the strengths and limitations of these approaches, providing empirical evidence to guide future technology selection and optimization in this domain.
\end{itemize}

\begin{figure}[t]
    \centering
    \includegraphics[width=\linewidth]{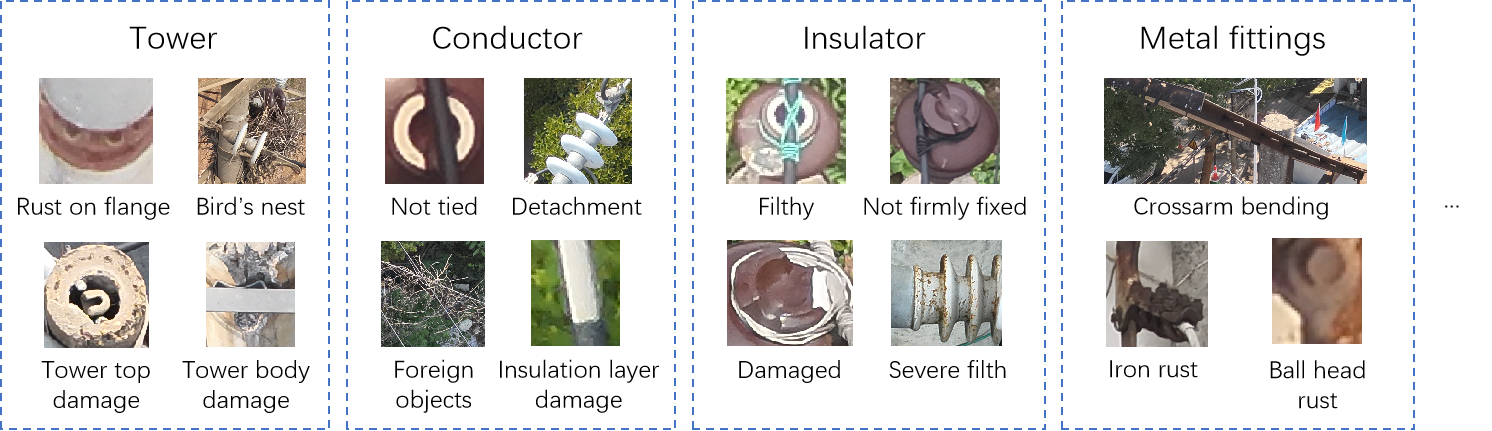}
    \caption{Representative examples of power equipment defects from the evaluation dataset. The images illustrate the diversity and visual complexity of anomalies across four primary component categories: towers, conductors, insulators, and metal fittings, highlighting the fine-grained perception capabilities required for automated inspection.}
    \label{fig:dataset_demo}
\end{figure}

\section{Related Work}

\subsection{Visual Language Model (VLM)}
Visual Language Models (VLMs) have significantly advanced artificial intelligence from perceptual to cognitive capabilities by unifying vision and language understanding. Pioneering models like CLIP \cite{radford_learning_2021} and ALIGN \cite{jia_scaling_2021} established zero-shot recognition through shared image-text semantic spaces, while subsequent architectures, including the BLIP series \cite{li_blip_2022, li2023blip}, InstructBLIP \cite{dai_instructblip_nodate}, and Flamingo \cite{alayrac_flamingo_2022}, enhanced multimodal fusion for complex tasks such as visual question answering. Recently, integrating powerful visual encoders with large language models has yielded large multimodal models (LMMs) capable of deep conversational interactions and complex reasoning, exemplified by LLaVA \cite{liu_visual_2023}, MiniGPT-4 \cite{zhu_minigpt-4_2023}, Qwen-VL \cite{wang2024qwen2}, and advanced proprietary models. Despite their demonstrated potential in areas like medical report generation \cite{li_llava-med_2023} and autonomous driving \cite{sima2024drivelm}, effectively applying VLMs to highly specialized power distribution inspection tasks remains a rarely explored frontier, particularly when the same model is expected to support fine-grained equipment recognition, defect description, severity grading, and maintenance-oriented decision support.

\subsection{Agents in Industrial Scenarios}
With the emergence of Large Language Models (LLMs), computing agents have evolved to utilize LLMs as their central ``brain'' for autonomous decision-making and goal execution \cite{xi_rise_2025, wang_survey_2024, park_generative_2023}. Leveraging robust reasoning capabilities and frameworks like ReAct \cite{yao_react_2022, wei_chain--thought_2022}, these LLM-driven agents decompose complex objectives into actionable subtasks and invoke external tools---ranging from APIs to physical actuators \cite{schick_toolformer_2023, qin_tool_2025}---to achieve closed-loop automation. While such agents have seen initial adoption in industrial automation \cite{xia_applying_2024, madani_large_2025, vyas2025autonomous}, our proposed framework conceptualizes an LLM-centric agent that ingests rich semantic inputs from VLMs to orchestrate power inspection workflows. By dynamically utilizing domain-specific toolkits, the agent connects visual defect understanding with knowledge-base retrieval, severity assessment, report generation, and alert/work-order actions, representing an exploration of how multimodal perception can be linked to operational maintenance decisions \cite{jin2025gridmind, wang2025llm}.

\subsection{Evaluation Tasks and Benchmarks}
Existing research on agent evaluation has established systematic multidimensional taxonomic frameworks, comprehensively extending testing scenarios from general common sense to complex physical and digital environments\cite{xie_large_2025,mohammadi_evaluation_2025,yehudai_survey_2025}. In terms of multimodal and expert-level cognition, the MMMU benchmark\cite{yue2024mmmu} challenges models' deep disciplinary perception and reasoning through over ten thousand university-level questions, while Ego-Exo4D\cite{grauman2024ego} leverages large-scale multi-view video datasets to focus on evaluating agents' recognition of fine-grained keysteps and skill proficiency. Regarding dynamic environmental interaction and complex workflow planning, GAIA\cite{mialon2023gaia} focuses on testing long-horizon planning capabilities in real-world multi-step tasks. Mind2Web\cite{deng2023mind2web} and its online version require agents to complete complex operations in authentic Web environments filled with interferences such as pop-ups and dynamic layouts, paying high attention to the correctness of the execution process. Furthermore, the recently proposed AgentVista\cite{su2026agentvista} challenges the long-horizon tool-invocation capabilities of generalist multimodal agents driven by visual evidence across multiple sub-domains.

\subsection{Evaluation Metric Systems}
Accompanied by the evolution of agent capabilities, evaluation metrics have developed from single lexical overlap measures (e.g., BLEU \cite{papineni2002bleu} and ROUGE \cite{lin2004rouge}) into a multidimensional system encompassing perception, behavior, reasoning, and system efficiency. On the foundational generation and visual perception fronts, researchers have introduced Pass@$k$ to measure functional correctness \cite{chen2021evaluating}, preference-based Elo ratings \cite{zheng2023judging}, as well as the POPE \cite{li2023evaluating} and CHAIR \cite{rohrbach2018object} frameworks specifically designed to detect and quantify visual hallucinations. For long-horizon tasks and dynamic interactions, the core of evaluation has shifted to Success Rate (SR) \cite{liu2023agentbench}, tool invocation, and retrieval accuracy (e.g., MRR and NDCG \cite{schwartz2021ensemble}), and the User-Sim Index (USI) \cite{zhou2026mind}, which quantifies the sim-to-real behavioral gap between simulators and real humans. Moreover, facing the challenge of sparse rewards in complex tasks, the evaluation focus is migrating towards the execution process: Process Reward Model (PRM) scores \cite{lightman2023let} and hindsight logic backtracking-based credit assignment frameworks (e.g., HCAPO \cite{tan2026hindsight}) are utilized to accurately assess the logical consistency and action necessity of intermediate reasoning steps. Meanwhile, efficiency metrics such as Time to First Token (TTFT), throughput, and per-task cost have become crucial standards for measuring the actual production deployment value of models \cite{liu2025speculative}.

\section{Method}
To systematically evaluate autonomous capabilities in power distribution inspection, we designed a comprehensive \textbf{Multi-Modal Agent} framework. Conceptually, this agent is defined as a computational entity capable of independently executing complex inspection tasks by closing the loop from perception to action. This section presents the overall design of the framework, the dataset, and the evaluation protocol.

\subsection{Overall Framework}
The agent's architecture is structured around three key components: the core cognitive engine, the interaction mechanism, and the configuration strategy.

(1) \textbf{The Foundation Model} is the heart of our agent. Rather than utilizing separate modules for vision and logic, we employ this single foundation model as the agent's core reasoning engine. This model is responsible for processing multi-modal information and driving the agent's three core capabilities: perception, reasoning, and tool usage.

(2) \textbf{Input and Output Mechanism.} The agent interacts with the environment through a streamlined multi-modal interface that processes a combined input stream of high-resolution inspection images and natural language task instructions. This dual-input capability allows the agent to analyze visual signals within the context of specific user queries. Conversely, the agent is designed to produce dual-modality outputs: it generates detailed, human-like natural language descriptions to explain equipment status and defects, while simultaneously producing machine-parsable structured commands in JSON format to specify tool calls and parameters for automated execution.

(3) \textbf{Prompt Configuration.} To adapt the general-purpose foundation model to the specialized power domain, we implement a rigorous prompt engineering strategy that steers the agent's behavior through a fixed template. This configuration explicitly defines the agent's role as an experienced power inspection expert to establish a professional context, strictly sets task constraints to valid equipment and defect categories to minimize hallucinations, and incorporates in-context learning via few-shot examples (comprising reference images and expert-verified descriptions) to standardize reasoning logic and output formats.

\subsection{Implementation of Three Core Capabilities}
We implement the agent's functionality by grounding the foundation model's behavior in three distinct capabilities, ensuring a robust transition from visual input to executable action.

\subsubsection{Perception}  The perception capability is responsible for extracting deep semantic information from visual signals. Unlike traditional detectors that output discrete labels, our implementation leverages the foundation model's vision-language alignment to: (1) Identify and Locate: Analyze the image content to identify equipment types and locate key components; (2) Semantic Description: Generate detailed, natural language descriptions of potential defects. For instance, rather than simply flagging an object as ``damaged,'' the model describes the specific visual features, such as ``rust on the flange''. This detailed semantic output bridges the gap between raw pixel data and expert-level understanding.

\subsubsection{Reasoning} The reasoning capability interprets perception results and plans necessary interventions. To mitigate the hallucination risks common in general-purpose models, we enhance this stage with standards-document Retrieval-Augmented Generation (RAG) using a domain knowledge base.
(1) \textbf{Knowledge retrieval}: The system retrieves relevant context from two sources: ``Power Equipment Defect Rating Standards'' for regulatory compliance and a ``Historical Defect Case Library'' for comparative analysis;
(2) \textbf{Logical inference}: The model combines the visual description with retrieved knowledge to perform comprehensive reasoning. For example, upon detecting flashover traces, the agent queries the standards to confirm that ``flashover'' constitutes a serious defect. It then plans the subsequent logic: rating the defect level as `critical', generating an emergency repair order, and archiving the report.

\subsubsection{Tool Usage} The tool usage capability enables the agent to interact with the external environment and close the inspection loop. The foundation model is instructed to function as an autonomous operator, selecting appropriate tools and generating precise arguments to execute tasks. We define a standardized set of API tools for the model to invoke:

\begin{itemize}
    \item ``get\_camera\_image(location)'': Retrieves the current camera feed or image data from a specified coordinate or device location.
    \item ``search\_knowledgebase(defect)'' Actively queries the internal database for solution protocols based on the identified defect information.
    \item ``write\_report(details)'': Compiles the analysis and reasoning results into a standardized inspection report format for archiving.
    \item ``send\_alert(message)'': Triggers an immediate notification to the relevant personnel in charge when high-risk defects are detected.
\end{itemize}

By strictly defining these interfaces, we evaluate the model's ability to not only ``think'' but to ``do''---accurately parsing user queries and generating correct function calls to drive the operation.

\section{Experiments}
We conduct a systematic and multi-dimensional performance evaluation of our proposed framework. The experiments are organized around three parts: first, we evaluate the basic capabilities of different visual language models (VLMs) and their performance under different prompting strategies; second, we explore the effect of exemplar retrieval on VLM performance; finally, we analyze the effectiveness of agents with different designs and cores in performing downstream automation tasks. All experiments are conducted on the private evaluation dataset.

\subsection{Experimental Setup}
To rigorously evaluate the proposed framework on power distribution inspection, we constructed a comprehensive multimodal dataset. The dataset comprises 26,803 meticulously curated high-resolution images collected from drone inspection and field inspection records over the past three years. It encompasses a wide range of geographical environments and climatic conditions, including sunny and cloudy weather, as well as urban, rural, plain, hilly, and mountainous scenarios, ensuring the authenticity and diversity required for evaluating equipment recognition, defect classification, severity grading, and agent-based maintenance workflows.

\begin{figure}[t]
    \centering
    \includegraphics[width=\linewidth]{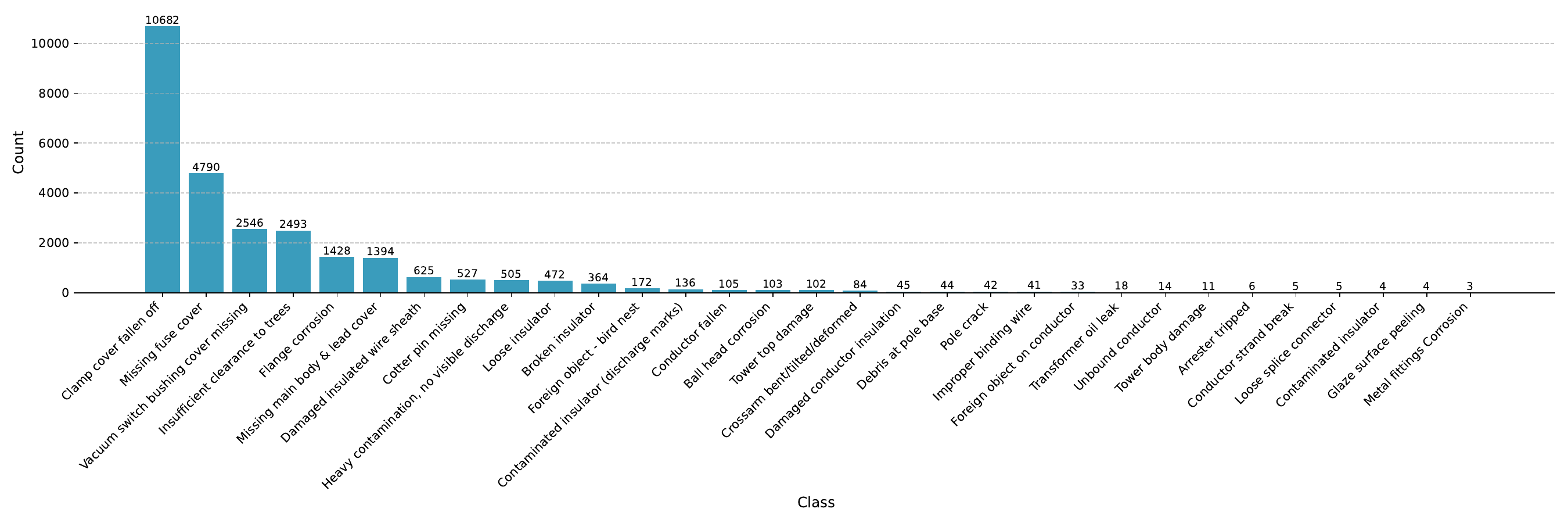}
    \caption{Distribution of defect categories in the dataset. The bar chart illustrates the highly imbalanced, long-tail nature of real-world power distribution defects.}
    \label{fig:dataset_dist}
\end{figure}

\subsubsection{Dataset Composition}
The dataset is organized with a fixed held-out evaluation split while preserving the long-tail distribution observed in real inspection records. It covers 10 equipment categories and 31 defect categories. For evaluation, each category contains no more than 2,000 held-out test samples. The dataset contains a portion of majority classes with abundant samples (representing the most common operational anomalies), alongside a significant long-tail portion consisting of rare, low-frequency categories. This natural imbalance is deliberately preserved to explicitly test the model's few-shot and zero-shot generalization capabilities when confronted with novel or scarce anomalies.

All data instances are manually annotated to support multidimensional perception and deep reasoning, with the following key elements:
\begin{itemize}
\item Image Data: The high-resolution visual input.
\item Category Information: Hierarchical labels detailing both the specific equipment type and the corresponding defect/anomaly category.
\item Textual Descriptions: Detailed natural language semantics capturing the visual context and anomaly characteristics.
\item Severity Levels: Graded assessments indicating the criticality or magnitude of the defect.
\item Spatial Location Information: Precise positional data and coordinates of the equipment for accurate spatial localization.
\end{itemize}
Furthermore, to enhance the model's expert-level cognitive and diagnostic abilities, we integrated an authoritative domain-specific defect regulation and standard protocol document. This document acts as an external knowledge base, utilized to drive Retrieval-Augmented Generation (RAG) processes. By retrieving highly relevant guidelines, it ensures that the agent's reasoning and final assessments strictly align with established industry standards.

To reduce evaluation ambiguity, the equipment and defect vocabularies used in prompts are fixed before inference and are kept consistent with the annotation schema. The few-shot/RAG exemplar pool contains at most 10 examples per category, is separated from the held-out evaluation images, and retrieved examples are selected only from this pool to avoid test-image leakage. Text-only RAG provides the exemplar description, equipment label, and defect label, whereas multimodal RAG additionally provides the corresponding reference image. Unless otherwise specified, all models use the same prompt template, the same exemplar budget for each X-shot setting, the same image inputs, and deterministic decoding settings when supported by the model implementation.

\subsubsection{Evaluation Metrics and Protocol}
To ensure reproducibility we follow a fixed evaluation protocol:
\begin{itemize}
    \item Detection / Recognition metrics: for the main recognition benchmark, \textit{Acc} is computed as a label-containing exact-match metric, i.e., a prediction is counted as correct when the generated text explicitly contains the ground-truth equipment or defect label. \textit{Recall}, \textit{Precision}, and \textit{F1} are reported as macro-averaged scores over classes after mapping unmatched predictions to an additional \texttt{other} class.
    \item Grading / Severity metrics: exact-match rate against expert-assigned grades.
    \item End-to-end task metrics: task success rate computed by comparing agent-generated actions (e.g., work orders) with ground-truth expected actions.
\end{itemize}

\subsection{Evaluation Details}
We evaluated a set of representative multimodal foundation models, including GLM-4.5V, Qwen2.5-VL-32B, Qwen3-VL-30B, LLaVA, DeepSeek-VL2, the Gemma3 family (4B, 12B, and 27B), and Step3. For reasoning-oriented evaluation, we also include models or configurations with explicit thinking capabilities, such as GLM-4.5V-thinking. Unless otherwise stated, all evaluations use deterministic decoding, the same prompt template, and the same held-out evaluation split. We only test locally deployable or data-security-compliant models to avoid exposing private inspection images.

To ensure reproducibility, we use a fixed prompt template and report all model results under the same zero-shot, 1-shot, and 5-shot settings. \textbf{(1) For zero-shot inference,}
this prompt directly instructs the model without providing any examples.

For example: \textit{You are an experienced power inspection expert. Please carefully analyze the following image, identify all power equipment in the image, and describe any defects in detail. Equipment component names include: [All equipment names]. If there are no defects, please answer ``No defects.'' If there are defects, please select the most appropriate defect type from the following categories: [Pre-defined defect names]. [Image].}

\textbf{(2) For few-shot inference}, additional reference sample information is provided at the beginning.

For example: \textit{Reference image: [image id], Similarity: [cosine similarity of embeddings], Description: [Specific appearance feature description]; Equipment type: [Equipment type]; Defect type: [Defect type].}

\begin{figure}[t]
    \centering
    \includegraphics[width=\linewidth]{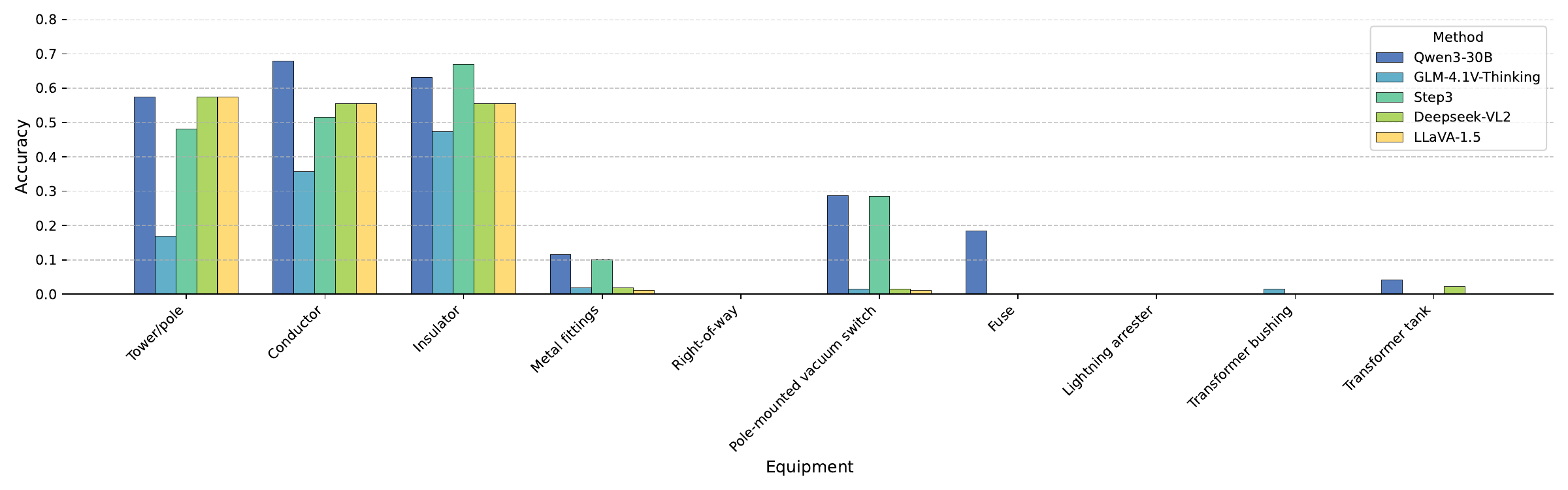}
    \caption{Accuracy of different multimodal models on equipment recognition across various power components.}
    \label{fig:equip1}
\end{figure}

\subsubsection{Recognition Performance}
In Table \ref{table:main}, we analyze the recognition capabilities of several mainstream existing multimodal large models for power equipment and defects. Here, \textit{Acc} denotes label-containing exact-match accuracy, while \textit{Recall}, \textit{Precision}, and \textit{F1} denote macro-averaged scores with unmatched generations mapped to an \texttt{other} class. Under this protocol, equipment recognition is consistently easier than defect recognition; however, the overall recognition performance remains unsatisfactory, with most zero-shot accuracy rates below 10\%. Additionally, Figure \ref{fig:equip1} presents the prediction accuracy of the models for each specific category. Specifically, all models exhibit generally higher recognition accuracy for basic objects (e.g., power poles, conductors, and insulators). Nevertheless, when confronted with more specialized power equipment (e.g., fuses and surge arresters), the performance of these models degrades significantly. This observation indicates that the pre-training data of general-purpose multimodal foundation models endows them with universal recognition capabilities for common objects, while they lack precise cognition of specific components within specialized industrial domains.

\begin{figure}[t]
    \centering
    \begin{subfigure}[b]{0.44\textwidth}
        \centering
        \includegraphics[width=\textwidth]{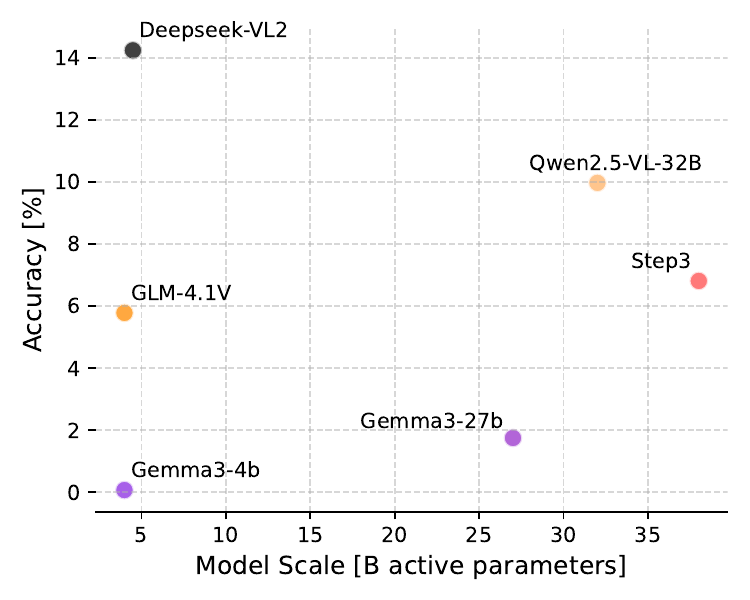}
        \caption{Equipment}
        \label{fig:sub1}
    \end{subfigure}
    \begin{subfigure}[b]{0.44\textwidth}
        \centering
        \includegraphics[width=\textwidth]{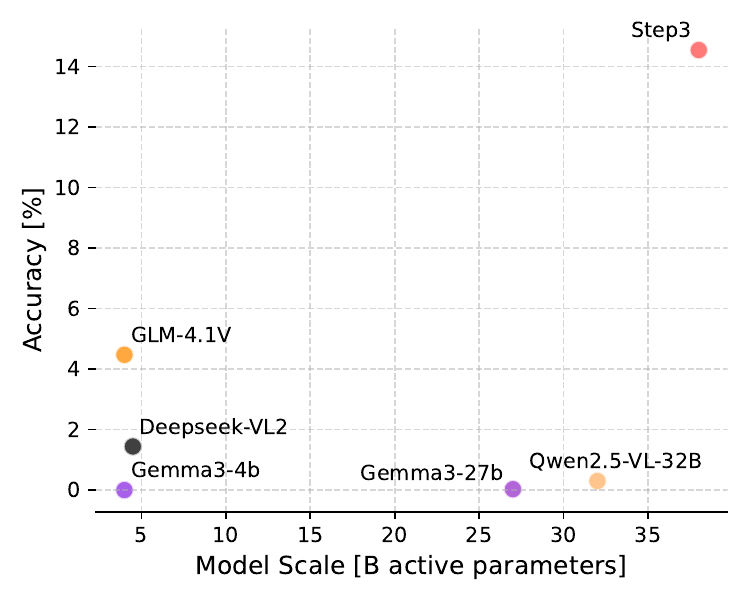}
        \caption{Defect}
        \label{fig:sub2}
    \end{subfigure}

    \begin{subfigure}[b]{0.44\textwidth}
        \centering
        \includegraphics[width=\textwidth]{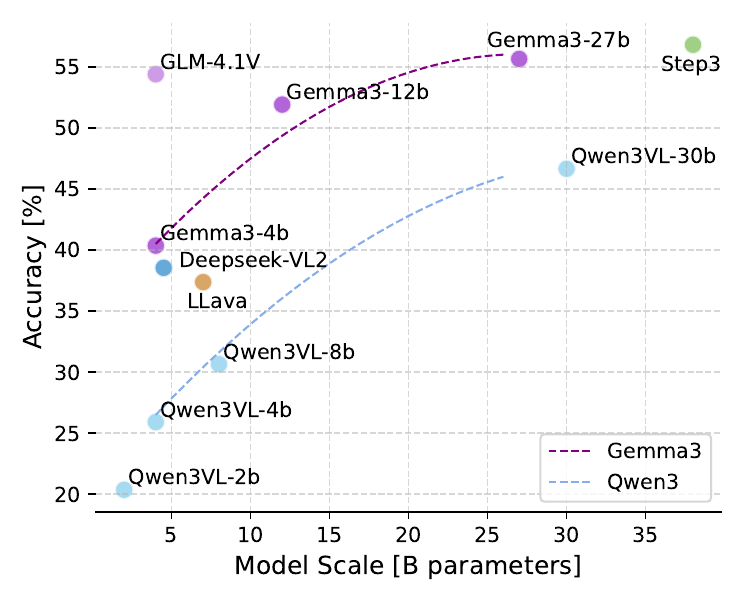}
        \caption{Equipment (5-shot RAG)}
        \label{fig:sub3}
    \end{subfigure}
    \begin{subfigure}[b]{0.44\textwidth}
        \centering
        \includegraphics[width=\textwidth]{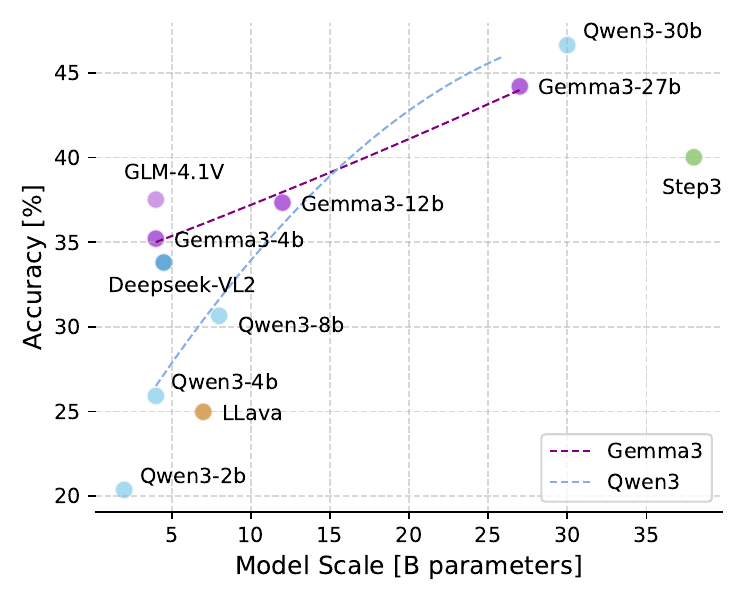}
        \caption{Defect (5-shot RAG)}
        \label{fig:sub4}
    \end{subfigure}
    \caption{Comparison of equipment and defect recognition accuracy across models of different scales, with and without RAG.}
    \label{fig:scale}
\end{figure}

\subsubsection{Impact of Model Scale}
As shown in Fig \ref{fig:scale}, for the power equipment and defect dataset focused on in this study, without the assistance of domain-related knowledge, the overall performance of the models is unsatisfactory, with the accuracy generally below 10\%; even when adopting model architectures with larger parameter scales, there is no significant improvement in performance. However, after integrating domain-specific knowledge of power equipment, the model performance has achieved substantial enhancement. At this point, there is an obvious positive correlation between model performance and parameter scale---models with larger parameter scales often demonstrate superior task processing capabilities.

\subsubsection{Effectiveness of Exemplar Retrieval}

Table \ref{table:main} presents a comparative analysis of model performance enhanced by exemplar retrieval, where the term ``X-shot'' denotes that X exemplars are supplemented for each category. Compared with the foundation-model performance without additional domain exemplars, exemplar retrieval significantly boosts recognition capabilities for both power equipment and defects. Furthermore, as shown in Figure \ref{fig:scale}, a positive correlation is observed between the number of retrieved exemplars and performance improvement; however, this enhancement is subject to a marginal effect. Specifically, when the number of supplementary exemplars reaches a certain threshold, the subsequent performance gain becomes negligible.

Two distinct exemplar-retrieval paradigms are further compared in Table \ref{table:mmrag}: one solely providing textual descriptive information of exemplars, and the other integrating both textual descriptions and corresponding image data of exemplars. Experimental results show a trade-off rather than a uniformly dominant choice. Textual descriptive information alone already yields a strong improvement and provides more stable gains for some accuracy-oriented metrics. Adding image information brings limited additional benefit overall: for DeepSeek under the 5-shot setting, equipment recognition accuracy increases by 2.05\%, while several equipment metrics decrease; for defect recognition, accuracy decreases by 0.64\%, but recall, precision, and F1 improve. This suggests that visual exemplars can help some recall-oriented defect metrics but may also introduce noise or distractors in fine-grained equipment recognition.

\begin{table}[!ht]
    \centering
    \caption{Performance of equipment and defect prediction under 5-shot settings using text-only and multimodal inputs. (\%)}
    \begin{tabular}{l|llll}
        \toprule
        Equipment & Acc & Recall & F1 score & Prec \\
        \midrule
        Descriptions (5-shot)  & 38.54 & 34.16 & 45.28 & 72.72 \\
        + images (5-shot)  & 40.59 $\uparrow$ & 32.86 $\downarrow$ & 42.13 $\downarrow$ & 63.63 $\downarrow$ \\
        \midrule
        Defect & Acc & Recall & F1 score & Prec \\
        \midrule
        Descriptions (5-shot)  & 33.80 & 40.98 & 51.73 & 81.48 \\
        + images (5-shot) & 33.16 $\downarrow$ & 45.96 $\uparrow$ & 58.19 $\uparrow$ & 88.88 $\uparrow$ \\ \bottomrule
    \end{tabular}
    \label{table:mmrag}
\end{table}

\begin{table}[tbp]
\centering
\caption{End-to-end grading accuracy and grading accuracy conditioned on correct defect prediction. (\%)}

\begin{tabular}{lrrrrr}
\toprule
Model                        & Step3 & Gemma3-27b & deepseek-vl2 & GLM-4.5V-thinking & Qwen3-VL-30B \\
\midrule
Grading Accuracy                 & 36.34 & 42.71      & 31.56        & 34.89             & 38.32        \\
Conditional Grading Accuracy     & 91.02 & 90.45      & 90.68        & 93.02             & 91.68       \\
\bottomrule
\end{tabular}
\label{table:grade}
\end{table}

\subsubsection{Defect Grading \& Analysis}
We observe that the agent's overall grading accuracy is limited, primarily due to errors propagated from the defect recognition stage. When conditioning the evaluation on samples with correctly predicted defects, however, the agent achieves substantially higher grading accuracy, as reported by the conditional grading accuracy in Table \ref{table:grade}. This indicates that the agent is generally capable of efficiently locating the relevant regulatory clauses and making correct judgments when provided with accurate defect information.

\begin{table}[tbp]
\caption{Quantitative comparison of baseline multi-modal models for equipment and defect recognition under zero-shot, 1-shot, and 5-shot configurations. Acc denotes label-containing exact-match accuracy. Recall, F1score, and Prec denote macro-averaged metrics with unmatched predictions mapped to an additional \texttt{other} class.}
\centering
\resizebox{\textwidth}{!}{%
\begin{tabular}{ll|rrrr|rrrr|rrrr}
\toprule
\multicolumn{2}{l|}{Model}       & \multicolumn{4}{c}{GLM-4.5V} & \multicolumn{4}{c}{Qwen2.5-vl-32B} & \multicolumn{4}{c}{Step3}        \\
\midrule
\multicolumn{2}{l|}{Metric}      & Acc     & Recall  & F1score  & Prec   & Acc    & Recall  & F1score & Prec  & Acc   & Recall & F1score & Prec  \\
\midrule
\multirow{2}{*}{0-shot} & equip  & 5.78    & 10.13   & 14.89    & 50     & 9.97   & 22.72   & 28.67   & 72.72 & 6.81  & 14.79  & 16.91   & 27.77 \\
                       & defect & 4.47    & 12.22   & 14.05    & 20.83  & 0.3    & 1.55    & 2.64    & 22.22 & 14.54 & 20.67  & 27.78   & 50    \\
\multirow{2}{*}{1-shot} & equip  & 45.32   & 39.43   & 49.88    & 72.72  & 24.91  & 26.26   & 36.68   & 72.72 & 46.52 & 41.72  & 50.96   & 72.72 \\
                       & defect & 32.85   & 50.64   & 60.15    & 88.88  & 1.27   & 6.06    & 8.04    & 55.55 & 34.88 & 51.05  & 60.34   & 88.88 \\
\multirow{2}{*}{5-shot} & equip  & 54.39   & 43.54   & 53.3     & 72.72  & 14.56  & 29.4    & 38.52   & 85.71 & 56.81 & 47.48  & 56.16   & 72.72 \\
                       & defect & 37.52   & 45.95   & 55.78    & 85.18  & 1.32   & 0.4     & 0.8     & 22.22 & 40.02 & 51.9   & 60.46   & 81.48 \\
\toprule
\multicolumn{2}{l|}{Model}       & \multicolumn{4}{c}{Gemma3-4b}         & \multicolumn{4}{c}{Gemma3-12b}     & \multicolumn{4}{c}{Gemma3-27b}   \\
\midrule
\multicolumn{2}{l|}{Metric}      & Acc     & Recall  & F1score  & Prec   & Acc    & Recall  & F1score & Prec  & Acc   & Recall & F1score & Prec  \\
\midrule
\multirow{2}{*}{0-shot} & equip  & 0.07  & 0.12  & 0.24  & 9.09  & 0.31  & 0.6   & 0.81  & 9.09  & 1.75  & 3.34  & 5.56  & 27.27 \\
                        & defect & 0     & 0     & 0     & 0     & 0     & 0     & 0     & 0     & 0.03  & 0.1   & 0.19  & 3.71  \\
\multirow{2}{*}{1-shot} & equip  & 30.75 & 32.32 & 41.99 & 72.72 & 33.67 & 36.28 & 43.65 & 72.72 & 36.85 & 40.67 & 46.58 & 72.72 \\
                        & defect & 14.08 & 27.45 & 35.71 & 77.77 & 27.52 & 33.46 & 39.96 & 77.77 & 33.36 & 37.48 & 44.12 & 66.66 \\
\multirow{2}{*}{5-shot} & equip  & 40.36 & 36.23 & 46.48 & 72.72 & 51.91 & 42.74 & 46.48 & 72.72 & 55.65 & 44.92 & 53.79 & 72.72 \\
                        & defect & 35.21 & 47.44 & 56.67 & 81.48 & 37.34 & 50.14 & 59.98 & 81.48 & 44.21 & 56.10 & 65.31 & 85.18 \\
                       \toprule
\multicolumn{2}{l|}{Model}        & \multicolumn{4}{c}{Qwen3-VL-30B}         & \multicolumn{4}{c}{Deepseek-vl2}     & \multicolumn{4}{c}{LLava}   \\
                       \midrule
\multicolumn{2}{l|}{Metric}       & Acc     & Recall  & F1score  & Prec   & Acc    & Recall  & F1score & Prec  & Acc   & Recall & F1score & Prec  \\
\midrule
\multirow{2}{*}{0-shot} & equip  & 4.12  & 6.98  & 9.55  & 54.54 & 14.24 & 15.84 & 22.23 & 54.54 & 11.84 & 12.65 & 20.45 & 54.54 \\
                       & defect & 0     & 0     & 0     & 0     & 1.44  & 3.01  & 4.84  & 33.36 & 1.44  & 2.56  & 3.68  & 33.36 \\
\multirow{2}{*}{1-shot} & equip  & 44.81 & 38.7  & 49.43 & 72.72 & 33.17 & 31.07 & 41.81 & 72.72 & 33.07 & 31.6  & 42.93 & 72.72 \\
                       & defect & 39.86 & 60.91 & 69.11 & 88.88 & 20.67 & 20.27 & 27.8  & 66.66 & 23.66 & 31.11 & 41.84 & 81.48 \\
\multirow{2}{*}{5-shot} & equip  & 46.72 & 39.73 & 49.78 & 72.72 & 38.54 & 34.16 & 45.28 & 72.72 & 37.37 & 32.96 & 44.57 & 72.72 \\
                       & defect & 40.74 & 60.75 & 69.76 & 88.88 & 33.8  & 40.98 & 51.73 & 81.48 & 24.97 & 25.48 & 35.35 & 77.77 \\
\hline
\end{tabular}%
}
\label{table:main}
\end{table}

\begin{figure}[t]
    \centering
    \includegraphics[width=\linewidth]{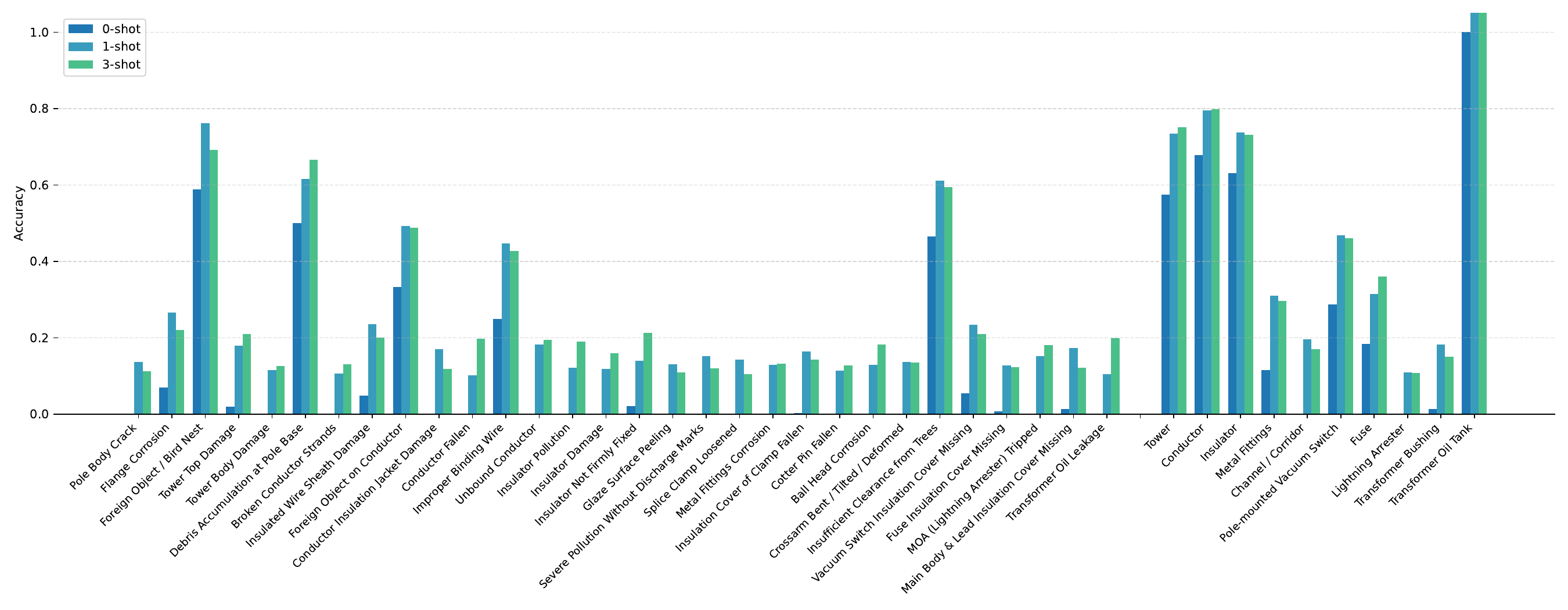}
    \caption{Per-class Detailed accuracy (Defect and equipment)}
    \label{fig:perclass}
\end{figure}

\begin{table}[tbp]
\caption{Quantitative assessment of tool usage and task execution metrics for various multi-modal large language models, categorized by their function-calling support.}
\centering
\begin{tabular}{l|lll|lll}

\toprule
\multirow{2}{*}{Models} & \multicolumn{3}{l}{Function call supported} & \multicolumn{3}{l}{Function call Not supported} \\
                        & Step3        & GLM-4.5V       & Qwen3-VL-30B       & Deepseek-vl2     & Qwen2.5-vl-32b     & Gemma3-12b    \\
                        \midrule
Tool usage accuracy     & 0.7083       & 0.7958         & 0.316       & 0.4429     & 0.6500             & 0.7100          \\
Argument accuracy       & 0.8042       & 0.7875         & 0.498       & 0.5762     & 0.9000             & 0.8600          \\
Toolchain coherence     & 0.6042       & 0.7708         & 0.305       & 0.3524     & 0.5375             & 0.5812        \\
Task success rate       & 0.4292       & 0.3000         & 0.034      & 0.1000    & 0.1833             & 0.2164       \\
\bottomrule
\end{tabular}
\label{table:tool_usage}
\end{table}

For tool-use evaluation, \textit{Tool usage accuracy} measures whether the model selects the correct tool at each step, averaged over all tool-selection steps. \textit{Argument accuracy} measures whether the arguments passed to the selected tools are correct, and the final score is averaged across the whole tool chain. \textit{Toolchain coherence} evaluates the correctness of the predicted tool chain against the ground-truth tool chain; we use an LLM judge to match the predicted and reference tool sequences and assign a score from 1 to 10, which is then normalized for reporting. \textit{Task success rate} measures end-to-end task completion: a task is counted as successful if the agent reaches the specified target, such as sending an alert or generating an inspection report, and unsuccessful otherwise.

In this section, we evaluate the execution performance of the agent implemented via the ReAct architecture. The evaluation focuses on the agent's ability to autonomously navigate the three-stage pipeline: Input Parsing, Task Decomposition, and Tool Invocation. Our experiments reveal distinct failure modes in both cognitive planning and operational execution, highlighting the gap between general-purpose capabilities and domain-specific requirements.

\subsubsection{Task Decomposition}
The primary bottleneck observed in the decomposition phase is the knowledge barrier. The task decomposition agent lacks internalized professional knowledge in the field of power inspection, which frequently results in unreasonable task division. Specifically, the agent tends to hallucinate non-existent targets; for instance, it may arbitrarily divide a single region (e.g., ``area A'') into multiple sub-areas without any logical basis, or generate inspection tasks for non-existent equipment points. These illogical decompositions lead to the generation of redundant or invalid sub-tasks, resulting in significant resource waste and confusion during the execution of the inspection workflow

\begin{figure}[t]
    \centering
    \includegraphics[width=0.9\linewidth]{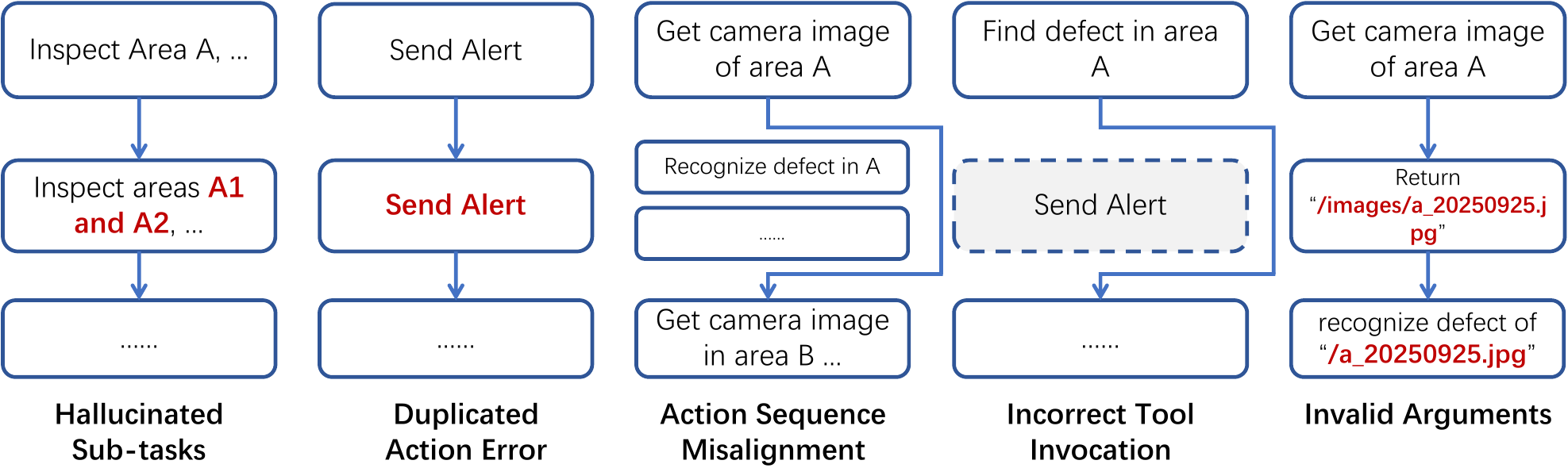}
    \caption{Typical categories of tool-execution errors in multimodal agents.}
    \label{fig:fail_case}
\end{figure}

\subsubsection{Tool Invocation}

In the execution phase, the agent struggles with toolchain coherence and cascading failures.
We observed that intelligent agents frequently misunderstand task intentions or tool capabilities, leading to disorganized calling sequences, such as circular or repeated tool calls. Furthermore, the workflow exhibits significant fragility; in a stepwise process, if an earlier step fails (e.g., failure to retrieve an image), subsequent steps often cannot run correctly due to missing prerequisites. This dependency failure triggers a chain reaction, resulting in inaccurate parameter passing, incoherent tool invocation chains, and a reduced overall task success rate, as illustrated in Fig. \ref{fig:fail_case}. These issues stem from the agent's insufficient understanding of tool usage constraints and the inherent instability of the fundamental model.

\section{Conclusion}

In this paper, we proposed a holistic evaluation framework for Multi-Modal Agents in power distribution inspection, treating them as unified cognitive engines capable of perception, reasoning, and tool usage. By establishing a domain-specific benchmark, we systematically assessed the readiness of fundamental models for autonomous industrial maintenance.
Our experiments reveal critical insights into the deployment of general-purpose Multi-Modal Large Language Models (MLLMs).

First, without domain adaptation, these models exhibit limited capability in recognizing fine-grained power equipment defects, achieving less than 10\% accuracy.

Second, while Retrieval-Augmented Generation (RAG) significantly bridges this semantic gap, we found that text-only and multimodal retrieval exhibit different trade-offs. Textual exemplars provide stable improvements and are often sufficient for accuracy-oriented recognition, whereas adding visual exemplars can improve some recall-oriented defect metrics but may also introduce noise for fine-grained equipment recognition.

Third, regarding agent autonomy, both cognitive planning and toolchain execution remain bottlenecks. Hallucinations during task decomposition frequently lead to invalid inspection targets, while cascading failures during tool invocation reduce end-to-end task success. Future work will focus on internalizing domain knowledge through instruction tuning to reduce dependence on external retrieval, and integrating advanced planning algorithms, such as self-reflection mechanisms, to enhance the robustness of agents in complex, long-horizon maintenance tasks.

\bibliographystyle{IEEEtran}
\bibliography{egbib}

\end{document}